# Interpretable Physics Extraction from Data for Linear Dynamical Systems using Lie Generator Networks

Shafayeth Jamil, and Rehan Kapadia
*Department of Electrical and Computer Engineering*
*University of Southern California*
sjamil@usc.edu, rkapadia@usc.edu

## Abstract

When the system is linear, why should learning be nonlinear? Linear dynamical systems, the analytical backbone of control theory, signal processing and circuit analysis, have exact closed-form solutions via the state transition matrix. Yet when system parameters must be inferred from data, recent neural approaches offer flexibility at the cost of physical guarantees: Neural ODEs provide flexible trajectory approximation but may violate physical invariants, while energy-preserving architectures do not natively represent dissipation essential to real-world systems. We introduce Lie Generator Networks (LGN), which learn a structured generator **A** and compute trajectories directly via matrix exponentiation. This shift from integration to exponentiation preserves structure by construction. By parameterizing **A** = **S** − **D** (skew-symmetric minus positive diagonal), stability and dissipation emerge from the underlying architecture and are not introduced during training via the loss function. LGN provides a unified framework for linear conservative, dissipative, and time-varying systems. On a 100-dimensional stable RLC ladder, standard derivative-based least-squares system identification can yield unstable eigenvalues. The unconstrained LGN yields stable but physically incorrect spectra, whereas LGN-SD recovers all 100 eigenvalues with over two orders of magnitude lower mean eigenvalue error than unconstrained alternatives. Critically, these eigenvalues reveal poles, natural frequencies, and damping ratios which are interpretable physics that black-box networks do not provide.

## Introduction

Learning dynamical systems from data is fundamental to science and engineering. Linear systems, described by $\dot{x}(t) = A(t)x(t)$ where matrix $A(t)$ is the system matrix, arise across domains, including circuit analysis, control theory, molecular dynamics, financial modeling, and linearization of nonlinear PDEs. When the generator $A(t)$ can be recovered from trajectory data, its eigenvalues yield modal structure, stability properties, characteristic timescales enabling not just prediction but interpretation and design. Classical system identification recovers $A(t)$ via least-squares regression on estimated derivatives [1]. This is well-posed in low dimensions with clean data, but scales poorly because derivative estimation amplifies noise, and unconstrained regression over n² parameters admits physically invalid solutions. Neural ODEs [2] elegantly parameterize continuous dynamics as $\dot{x}(t) = f_\theta(x, t)$ enabling learning from irregularly-sampled data. On the other hand energy-preserving neural networks such as HNNs [3], LNNs [4], and SympNets [5] impose structure by learning energy functions and preserving phase-space volume. However, these neural models share fundamental limitations. First, numerical integration accumulates error; structure preservation is approximate and degrades over long horizons. Second, energy-conserving methods in their standard form represent only conservative systems and dissipative extensions [9, 10] rely on numerical integration, inheriting its error accumulation. Third, the learned representations are opaque as a Neural ODE provides no direct identification of underlying system physics, such as poles, damping ratios, or natural frequencies. Fourth, for linear systems, these methods solve a harder problem than necessary, which is approximating ẋ = Ax with a nonlinear $f_\theta$ when the exact solution x(t) = exp(At)x₀ is available in closed form.

We introduce Lie Generator Networks (LGN), which learn the system matrix A and evolve states via matrix exponentiation rather than numerical integration. For the Linear time invariant (LTI) case, x(t) = exp(At)x₀ is evaluated directly. In the Linear time-varying (LTV) case, we employ the Magnus expansion [6] to handle non-commuting generators. This replaces derivative estimation and step-by-step integration with a structured generator-learning problem. The generator matrix exponential preserves the underlying structure. For example, if we constrain A to have certain properties such as eigenvalues with negative real parts, the solution exp(A)x₀ inherits those properties exactly.

LGN provides a unified framework for linear conservative, dissipative, LTI and LTV systems. By parameterizing **A** = **S** − **D**, where **S** is skew-symmetric and **D** is positive diagonal, the S-D decomposition guarantees all eigenvalues satisfy Re(λ) ≤ 0 and quadratic energy dissipates monotonically. This gives us structure preservation to machine precision that is enforced by the architecture. For LTV systems, the Magnus truncation does not give us machine precision, but inherits $O(\Delta t^{2m+1})$ local error as shown by classical Magnus theory [7]. This gives us rigorous error bounds. The S-D constraint also acts as implicit regularization, enabling robust eigenvalue recovery even from noisy measurements. Essentially, the learned system matrix, **A** directly reveals the physics of the underlying system, something other current machine system poles, natural frequencies, and damping ratios which are interpretable physics that black-box methods do not offer.

## Method

We study learning linear dynamical systems from trajectory observations. The systems under consideration here are based on data samples $\{(t_i, x_i)\}_{i=0}^{T}$ from an underlying system,
$$\dot{x}(t) = A(t) x(t).$$
The goal is to recover a generator $A(t)$ such that the learned dynamics match the data while preserving known physical structure (e.g., conservation, stability, dissipation). Rather than learning a generic nonlinear vector field and numerically integrating it, Lie Generator Networks (LGN) learn the generator directly and evolve states through matrix exponentiation, yielding closed-form flow maps for linear systems.

### LGN for LTI systems

For the linear time-invariant (LTI) case $A(t) \equiv A$, trajectories admit the closed form
$$x(t) = \exp(At) x_0.$$
LGN learns a structured parameterization $A_\theta$ and predicts
$$\hat{x}(t_i) = \exp(A_\theta t_i) x_0.$$
This approach can be described as learn $A$, then exponentiate, and avoids derivative estimation and numerical integration drift. Moreover, by constraining $A_\theta$ to a physically valid class, then $\exp(A_\theta t)$ inherits the corresponding structure by construction.

### LGN for LTV systems

For linear time-varying (LTV) dynamics, the solution can be written as a single exponential
$$x(t) = \exp(\Omega(t)) x_0,$$
where $\Omega(t)$ is given by the Magnus expansion

$$\Omega(t) = \int_0^t A(\tau_1)\,d\tau_1 - \frac{1}{2}\int_0^t d\tau_1 \int_0^{\tau_1} [A(\tau_1), A(\tau_2)]\,d\tau_2 + \cdots,\ [X,Y] = XY - YX.$$

The first term captures the accumulated generator, while higher-order terms correct for non-commutativity when $[A(\tau_1), A(\tau_2)] \neq 0$. For slowly varying systems or near-commuting generators, truncating to first order is often sufficient. When $A(t)$ varies more rapidly, we include the second-order commutator correction.

We roll out over discrete observation times $\{t_n\}_{n=0}^T$. Over each interval $[t_n, t_{n+1}]$, we approximate the transition matrix by a single exponential

$$\Phi(t_{n+1}, t_n) \approx \exp(\Omega_n^{(m)}),\ \hat{x}_{n+1} = \exp(\Omega_n^{(m)})\hat{x}_n,$$

where the superscript $m$ denotes Magnus truncation order (LGN-M$m$) and the subscript $n$ denotes the time-step index. Let $\Delta t_n = t_{n+1} - t_n$ and $t_{n+\frac{1}{2}} = (t_n + t_{n+1})/2$. Using midpoint quadrature, the first-order Magnus update is

$$\Omega_n^{(1)} \approx A(t_{n+\frac{1}{2}})\,\Delta t_n.$$

Including the second-order commutator correction yields the discrete Magnus-2 approximation used in this work:

$$\Omega_n^{(2)} \approx A(t_{n+\frac{1}{2}})\,\Delta t_n + \frac{\Delta t_n^2}{12}[A(t_n), A(t_{n+1})]$$

We refer to the $m$-th order truncation as LGN-M$m$ for LTV systems.

Under standard Magnus convergence assumptions, the $m$-th order truncation attains local error $O(\Delta t^{2m+1})$. Thus LGN-M1 and LGN-M2 yield $O(\Delta t^3)$ and $O(\Delta t^5)$ local error, respectively.

**Structure by construction: The $S - D$ generator decomposition**

To structurally enforce stability a, we parameterize the generator as

$$A = S - D,\ S = -S^\top,\ D = \text{diag}(d_i),\ \ d_i \geq 0.$$

This decomposition reflects the physical structure of dissipative systems: S captures conservative energy exchange between state variables, while D represents irreversible losses. We refer to this constrained model as LGN-SD. The unconstrained model that learns a full matrix $A$ is denoted LGN-FA.

This parameterization guarantees left-half-plane spectrum and monotonic decay of the quadratic energy $\| x \|^2$. Using the Lyapunov function $V(x) = \| x \|^2$,

$$\frac{d}{dt}\| x \|^2 = x^\top(A + A^\top)x = x^\top((S-D) + (S^\top - D))x = -2x^\top D x \leq 0.$$

Consequently, $\| x(t) \|^2$ decays monotonically and $\text{Re}(\lambda) \leq 0$ for all eigenvalues of $A$, ensuring dissipative stability by construction. Because LGN propagates with $\exp(\cdot)$, these guarantees hold at the level of the learned flow map.

**Learning objective and optimization**

Given trajectory data $\{(t_i, x_i)\}$, we minimize a reconstruction loss between observed states and LGN predictions:

$$\mathcal{L}(\theta) = \sum_i \| x_i - \hat{x}(t_i; \theta) \|^2 = \sum_i \| x_i - \exp(\Omega_\theta(t_i)) x_0 \|^2,$$

where $\Omega_\theta(t_i) = A_\theta t_i$ in the LTI case, and $\Omega_\theta(t_i)$ is obtained by composing per-step Magnus updates $\Omega_n^{(m)}$ over the observation grid in the LTV case. Gradients are computed by automatic differentiation through the matrix exponential (e.g., torch.matrix_exp), enabling end-to-end training of generator parameters and any time-parameterization used for $A_\theta(t)$.

# Experiments

We evaluate LGN on four systems of increasing complexity: conservative circuits (LC), dissipative circuits (RLC), time-varying dynamics (LTV oscillator), and higher-dimensional systems (100D RLC ladder). Each experiment isolates a distinct capability: (i) exact recovery when the hypothesis class matches the physics, (ii) modeling dissipation with stability guarantees, (iii) accounting for time-ordering via the Magnus commutator correction, and (iv) scalability to $n = 100$ with *interpretable* recovered spectra. We design four experiments, each testing these claims:

| Experiment | System | State Dimension | LGN learnable parameters |
|---|---|---|---|
| Exp 1: LC/RLC Circuit | Hamiltonian/dissipative | 2 | 1 (ω)/ 2 (ω, γ) |
| Exp 2: LTV Oscillator | Time-varying A(t) | 2 | 153 |
| Exp 3: RLC Ladder | 100-dimensional | 100 | 5,050 |
| Exp 4: Noise Robustness | 6-dimensional | 6 | 21 |

We compare against three baselines representing common identification strategies. Hamiltonian Neural Networks (HNN) learn a scalar energy function $H_\theta(q, p)$ using a two-hidden-layer MLP (64 units each, tanh activation, ~4,400 parameters) and derive dynamics through automatic differentiation: $\dot{q} = \partial H/\partial p$, $\dot{p} = -\partial H/\partial q$. Neural ODE uses an identical architecture but directly outputs the time derivative $\dot{x} = f_\theta(x)$ without structural constraints. Both HNN and Neural ODE are rolled out with an adaptive Runge–Kutta solver (tolerance $10^{-5}$). Linear System Identification (linear-ID) fits constant system matrix A via least squares on central-difference derivative estimates. This classical baseline requires no iterative training but is sensitive to noise and derivative estimation error. All neural models use Adam with learning rate, lr = $10^{-3}$ with learning rate scheduling (patience 200, factor 0.5), gradient clipping at 1.0, and train for 3000 epochs. Weights are initialized with Xavier normal (gain 0.5 for HNN, 0.3 for Neural ODE). For the 100D ladder (Experiment 3), we train for 1000 epochs due to computational cost. LGN uses lr = $10^{-2}$ due to low training parameter count. We report trajectory error Normalized RMSE (NRMSE) over the full rollout. For experiments where physical energy behavior is known, we additionally report an energy-increase violation rate:

$$\text{Viol}(E) = \frac{1}{T-1} \sum_{k=0}^{T-2} \mathbf{1}\{E_{k+1} > E_k + 10^{-6}\},$$

measuring the fraction of timesteps where energy increases. We define Normalized RMSE as NRMSE = RMSE / RMS, where $RMSE = \sqrt{\left(\frac{1}{T}\sum_i ||\hat{x_i} - x_i||^2\right)}$ and $RMS = \sqrt{\left(\frac{1}{T}\sum_i ||x_i||^2\right)}$ is the root-mean-square amplitude of the ground truth trajectory.

## Experiment 1: LTI Conservative and Dissipative systems

The simplest test of structure-preserving learning is to observe if a single framework can handle both energy-conserving and energy-dissipating dynamics. We study two canonical systems, the LC oscillator (conservative) and RLC oscillator (dissipative), which differ only in whether dissipation is present.

Both systems evolve as $\dot{x}$ = Ax with state x = $[q, p]^T$. The LC oscillator has generator

$$A = \begin{pmatrix} 0 & \omega \\ -\omega & 0 \end{pmatrix}$$

which is purely skew-symmetric (A = S, D = 0). The Hamiltonian $H(q,p) = ½(q^2 + p^2)$ is exactly conserved, and trajectories are rotations in phase space. (Fig 1(a))

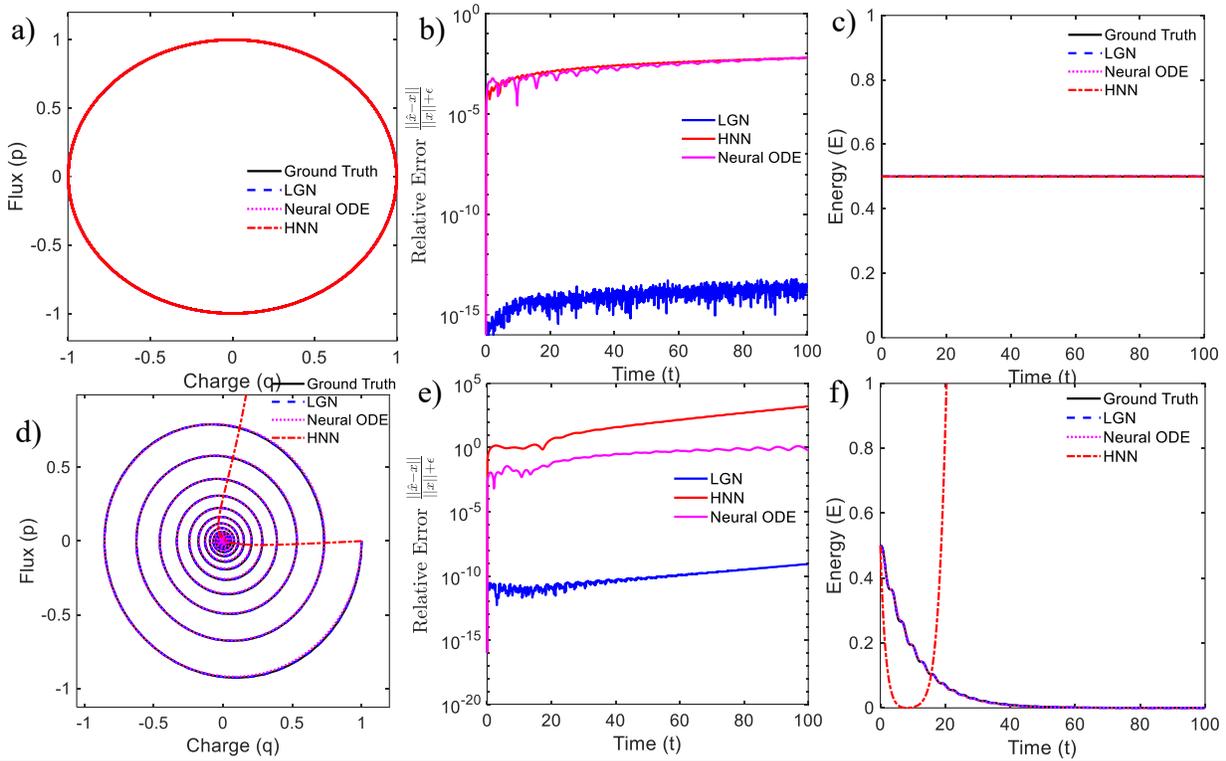

Figure 1: **LTI Systems**. Top row: Conservative/Hamiltonian (LC). Bottom row: Dissipative (RLC). (a,d) Phase portraits. (b,e) Relative error. (c,f) Energy. LGN achieves machine precision in both cases. HNN, constrained to symplectic structure, cannot model dissipation.

Adding resistance yields the RLC oscillator:

$$A = \begin{pmatrix} 0 & 1 \\ -\omega^2 & -\gamma \end{pmatrix}$$

where $\gamma > 0$ is the damping coefficient. With $H = \frac{1}{2}(\omega^2 q^2 + p^2)$, energy dissipates as $\dot{E} = -\gamma p^2 \leq 0$ for any $\omega$ ( $\omega = 1$ used here), $H = \frac{1}{2}(q^2 + p^2)$. Trajectories spiral inward to the origin. (Fig 1(d)).

**Setup**

For LC: $\omega = 1.0$, $x(0) = [1, 0]^T$. For RLC: $\omega = 1.0$, $\gamma = 0.1$, $x(0) = [1, 0]^T$. With $\omega = 1$, the RLC generator is exactly in S−D form ($S_{12} = -S_{21} = 1$, $D_{22} = \gamma$), and $H = \frac{1}{2}(q^2 + p^2)$ is the physical energy. Similarly, in Experiment 3, setting $L = C = 1$ places the ladder in energy coordinates where the Euclidean norm equals stored energy, making S−D representation exact. For general $\omega \neq 1$ or unequal L, C, a non-identity Lyapunov metric P would be required (see Limitations). Training uses $t \in [0, 10]$ for LC and $t \in [0, 20]$ for RLC. Both are tested on $t \in [0, 100]$ (5-10× extrapolation).

LGN learns only the physical parameters: ω for LC and $(\omega^2, \gamma)$ for RLC. HNN and Neural ODE use identical two-hidden-layer MLPs (64 units, tanh, ~4,400 parameters) representing standard practice for unknown nonlinear systems.

**Results**

| System | Method | NRMSE | Energy Metric |
|---|---|---|---|
| LC | LGN | $1.6\times10^{-14}$ | $\sigma_E = 1.3\times10^{-14}$ |
| LC | HNN | $3.6\times10^{-3}$ | $\sigma_E = 6.8\times10^{-5}$ |
| LC | Neural ODE | $3.3\times10^{-3}$ | $\sigma_E = 3.7\times10^{-4}$ |
| RLC | LGN | $2\times10^{-11}$ | 0.0% violation |
| RLC | HNN | $2.1\times10^{+1}$ | 92% violation |
| RLC | Neural ODE | $7.9\times10^{-2}$ | 14.5% violation |

LGN achieves machine precision in both cases because these experiments lie exactly in the Lie hypothesis class, learning ω or $(\omega^2, \gamma)$ identifies the generator, and matrix exponentiation provides the closed-form solution. The 12-order-of-magnitude gap versus neural baselines reflects a the difference in parameter identification versus function approximation.

The RLC results expose a structural limitation of Hamiltonian methods. HNN preserves its learned Hamiltonian by construction and dissipation is architecturally excluded. On LC this is correct behavior; on RLC it is catastrophic (NRMSE > 20, and 92% energy violations). This is not a training failure but a model class mismatch as symplectic structure cannot represent non-conservative dynamics. Neural ODEs are unconstrained by structure and therefore fit both systems to ~$10^{-2}$ accuracy. It can represent dissipation but provides no guarantees, as shown that 14.5% of RLC timesteps show energy increases that violate the underlying system physics.

The parameter disparity (2 vs 4,400) is due to the underlying enforced structure. When the dynamics are linear, LGN learns the physical parameters directly while neural methods must discover structure through function approximation. Figure 1 shows that this gap manifests in phase space: LGN trajectories overlay ground truth exactly, while neural predictions accumulate drift. This is where the underlying advantage of matrix exponentiation vs time-stepping methods.

# Experiment 2: LTV Dissipative (Time-Varying System)

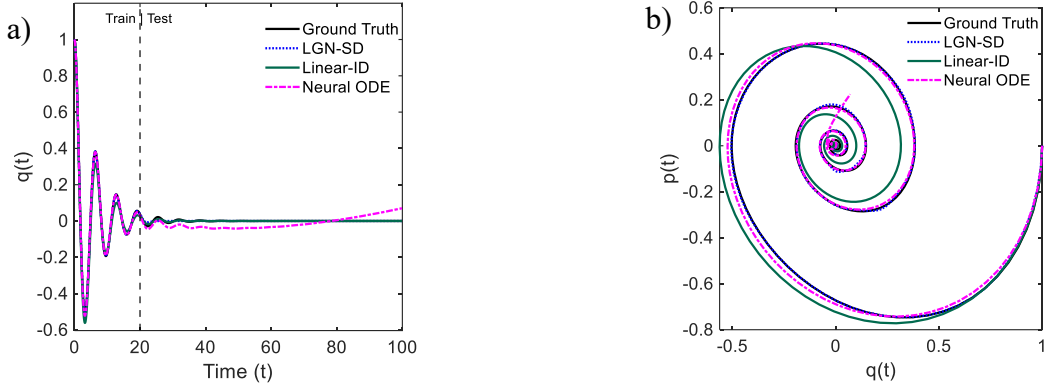

Figure 2. **2D Linear Time Variant Dissipative Systems.** a) The transient response; b) The phase space.

We consider a linear time-varying (LTV) oscillator where dissipation changes over time:

$$\frac{d}{dt}\begin{pmatrix}q\\p\end{pmatrix} = \begin{pmatrix} 0 & 1 \\ -\omega_0^2 & -\gamma(t) \end{pmatrix}\begin{pmatrix}q\\p\end{pmatrix}$$

where $\gamma(t) = \gamma_0(1 + \gamma_{\text{amp}}\sin(\omega_d t))$ varies continuously and is unknown to all models. Unlike LTI systems, LGN parameterizes the generator as $A(t) = W\phi(t) + b$. With scalar time ($\phi(t) = t$), this reduces to $A(t) = Wt + b$, which is a linear ramp that cannot capture oscillatory variation. We therefore sample 25 frequencies $\{\omega_k\}$ log-uniformly from $[0.10, 10.0]$ Hz; a broad range not tuned to the true modulation frequency $\omega_d = 1$. The time embedding is:

$$\phi(t) = [\cos(\omega_1 t), \sin(\omega_1 t), \ldots, \cos(\omega_{25} t), \sin(\omega_{25} t)] \in \mathbb{R}^{50}$$

This is a universal basis as any smooth periodic function can be approximated by linear combinations of these features. We parameterize the generator as $A(t) = S(t) - D(t)$ where: $S(t) = W_S \phi(t)$, and $D(t) = \text{diag}(\text{softplus}(W_D \phi(t)))$. This guarantees $\text{Re}(\lambda) \leq 0$ for all eigenvalues at all times, enforcing stability. The model learns $W_S \in \mathbb{R}^{1\times 50}$ and $W_D \in \mathbb{R}^{2\times 50}$, totaling 153 parameters. This approach is compared to a neural ODE. Since a neural ODE can learn arbitrary time-dependence through its MLP, we considered a reference neural ODE (Neural ODE-REF) that also receives the same features:

$$\dot{x} = f_\theta(x, \phi(t)), f_\theta: \mathbb{R}^{2+50} \to \mathbb{R}^2$$

This ensures both models have identical temporal information. Any performance gap is due to architecture, not time encoding.

**Setup**

We fix $\omega_0 = 1$, $\gamma_0 = 0.3$, $\gamma_{amp} = 0.15$, $\omega_d = 1$, $x(0) = [1,0]^T$. Training uses $t \in [0,20]$ with $\Delta t = 0.1$; testing uses $t \in [0,100]$ with $\Delta t = 0.1$.

**Results**

| Method | Parameters | NRMSE |
|---|---|---|
| LGN-SD (Frequency data) | 153 | 0.037 |
| Linear-ID | 4 | 0.11 |
| Neural ODE (time data) | ~4500 | 0.55 |
| Neural ODE (Frequency data) | ~7600 | 5.79 |
| Neural ODE (Frequency data) | 222 | 7.29 |

LGN-SD achieves NRMSE 0.037 with 153 parameters. As the commutator value is small (mean 0.013) adding commutator term in the magnus expansion does not yield improvement. In the appendix we show that when the underlying time varying dissipation equation is given, commutator inclusion improves the accuracy. When the parametric form of A(t) is known, LGN recovers physical parameters directly; with flexible S-D parameterization, LGN achieves trajectory accuracy without explicit parameter recovery in LTV systems. Neural ODE with scalar time (standard baseline) reaches NRMSE of 0.554; 15x worse despite 30x more parameters compared to LGN-SD. The last two rows isolate the role of time encoding, giving Neural ODE identical Fourier features degrades rather than improves performance (NRMSE 5.79 and 7.29), confirming that LGN's advantage stems from its linear generator inductive bias, not the temporal representation. Fourier features are necessary for LGN's linear readout to express smooth time-dependence, whereas NODE's MLP can learn this internally from scalar $t$. However, even with this apparent advantage, Neural ODE's unconstrained hypothesis class leads to overfitting. The linear inductive bias, not the time encoding, accounts for the 15-197× improvement. We also tested LGN without the S-D decomposition (LGN-FA), and this model diverged, confirming that stability structure is essential.

# Experiment 3: RLC Ladder (scaling to 100D)

We now test all methods on a 50-section RLC ladder network (100 state variables), representative of high-dimensional dissipative systems arising in thermal networks, mechanical systems with damping, and discretized PDEs. We set L = C = 1 and R = 0.1 for all sections, producing a system with eigenvalues clustered near $\text{Re}(\lambda) \approx -0.05$ with range $[-0.089, -0.011]$ due to boundary effects in the coupled oscillator chain. Training uses 3 trajectories with different initial conditions over t ∈ [0, 30]; testing extrapolates to t ∈ [0, 100]. We compare three approaches: Linear-ID (least-squares on estimated derivatives), LGN-FA (unconstrained n² parameters), and LGN-SD (S−D parameterization with $n(n-1)/2 + n \approx 5{,}050$ parameters). We omit Neural ODE for this experiment because it cannot recover the system matrix A; eigenvalue comparison is therefore impossible.

**Results**

| Method | NRMSE | Energy Violation | Mean $\text{Re}(\lambda)$ | Unstable Eigenvalues |
|---|---|---|---|---|
| Ground Truth | - | - | -0.050 | 0 |
| Linear-ID | $4.4 \times 10^{+54}$ | 87.7% | -0.018 | 16 |
| LGN-FA | 0.93 | 2.9% | -0.55 | 0 |
| LGN-SD | 0.90 | 0.0% | -0.051 | 0 |

Linear-ID fails catastrophically as of the 100 learned eigenvalues, 16 have positive real parts; the learned system is unstable and predicts exponentially diverging trajectories. Test NRMSE is effectively infinite ($4.4 \times 10^{54}$). This is not a hyperparameter issue; unconstrained least-squares on n² parameters is fundamentally ill-posed at this scale. LGN-FA finds a stable solution but does not learn the systems underlying physics. However, the mean eigenvalue real part is −0.55 versus the true mean of −0.05, yielding 500× larger error than LGN-SD. The optimizer has found a local minimum corresponding to a heavily overdamped system that fits training trajectories but misidentifies the underlying dynamics. In contrast, LGN-SD recovers the correct dynamics. Mean $\text{Re}(\lambda) = -0.051$ matches truth to within 2%. Fig. 3(b) shows eigenvalue error distributions where LGN-SD concentrates near zero (median < 0.01), while Linear-ID and LGN-FA show systematic biases with median errors of 0.05 and 0.39 respectively (~40× improvement). This experiment demonstrates that structural constraints are not optional regularization but necessary conditions for correct identification at scale. The S-D parameterization encodes the system class (i.e. dissipative) while still learning the system content, which are the eigenvalues. This is the gap between a predictive surrogate and a system identification tool. LGN-SD recovers poles, natural frequencies, and damping ratios that enable extrapolation, inverse design, and physical validation.

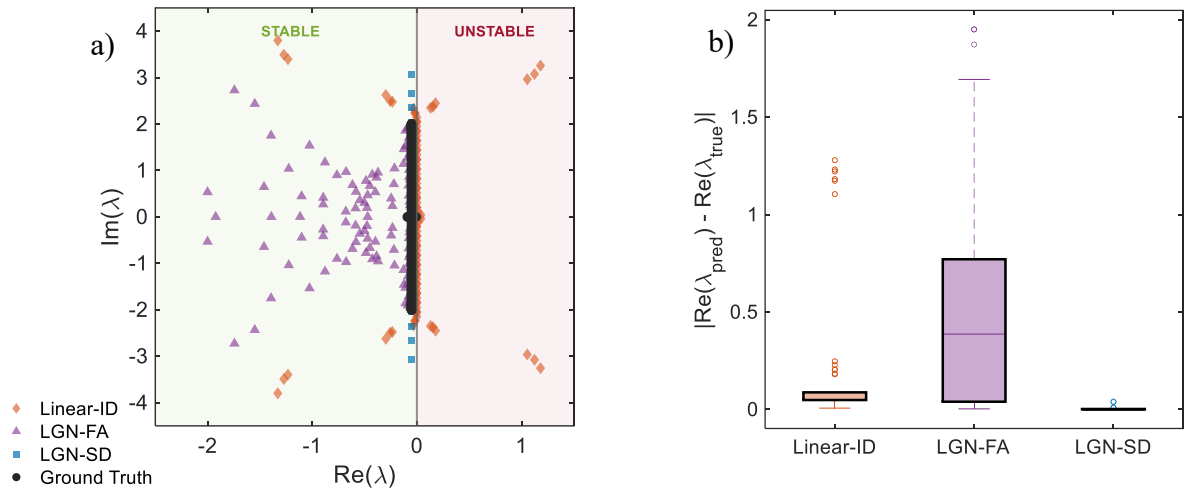

Figure 3. **High-dimensional system identification reveals the critical role of structural constraints.** (a) Eigenvalue spectrum in the complex plane. Ground truth eigenvalues (black dots) cluster at $Re(\lambda) \approx -0.05$. Linear System ID (orange) produces eigenvalues in the unstable half-plane ($Re(\lambda) > 0$). LGN-FA learning (purple) finds stable but incorrect eigenvalues. Only LGN-SD (cyan) recovers the true spectrum. (b) Distribution of eigenvalue recovery error $|Re(\lambda\_pred) - Re(\lambda\_true)|$. LGN-SD achieves near-zero error while alternatives exhibit systematic biases.

## Exp 4: Noise Robustness

Real measurements are never clean. Sensor noise, quantization error, and environmental interference are superimposed every trajectory collected by physical sensors. We evaluate eigenvalue recovery under additive Gaussian noise at 1%, 5%, and 10% of signal amplitude. Figure 4 shows that Linear-ID, which fits **A** via least-squares on numerical derivatives, degrades linearly with noise because derivative estimation amplifies noise. At 10% noise, eigenvalue error reaches ~ 8%. In contrast, LGN remains flat at <1% mean error across all noise levels. This result is due to the S-D parameterization constraining the solution to the manifold of stable generators. Regardless of the noise added to the trajectory samples in this experiment, the learned A must satisfy $Re(\lambda) \leq 0$. This structural constraint acts as implicit regularization, rejecting the high-variance solutions that are allowed in unconstrained fitting. In Appendix A.2, we further show that LGN-SD outperforms Dissipative SymODEN [8] by two orders of magnitude on stiff systems, where τ-horizon training fails to capture slow eigenvalues that LGN's full-trajectory matrix exponential directly recovers. This robustness has direct practical consequences as Eigenvalues encode stability margins, damping ratios, and resonant frequencies. These are quantities that determine the real world behavior of dynamical systems, and accurate recovery from noisy data enables reliable analysis even under real-world conditions.

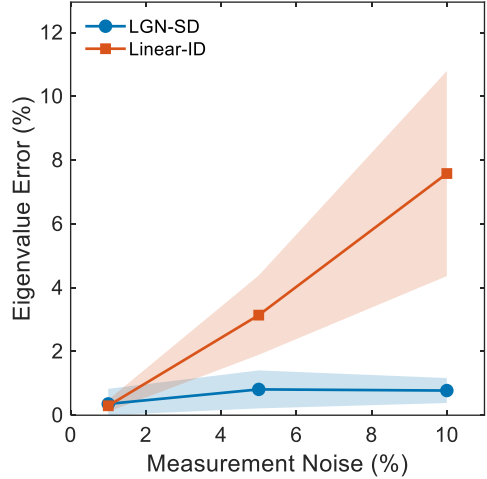

Figure 4. **Noise Robustness.** For 6D RLC system, the mean eigenvalue error remains below 1% as the noise in the training data grows to 10%. For Linear-ID the eigenvalue error reaches ~8% at 10% noise error.

## Related Work

**Neural ODEs and structure-preserving networks.** Neural ODEs parameterize dynamics as $\dot{x} = f_\theta(x, t)$ and backpropagate through numerical solvers. Hamiltonian Neural Networks learn scalar energy functions and derive dynamics via symplectic gradients, with extensions to Lagrangian and symplectic recurrent forms. Port-Hamiltonian Neural Networks handle dissipation by learning separate dissipation matrices [8]. Dissipative HNNs [9] parameterize a Rayleigh dissipation function alongside the Hamiltonian; and recent theoretical work establishes learnability conditions for linear port-Hamiltonian systems [10]. However, as we demonstrate in Appendix A.2, τ-horizon training creates an information bottleneck for stiff systems where slow modes are indistinguishable from constant offsets within short training windows. All these methods rely on numerical integration which makes structure preservation approximate and degrades over long horizons. LGN computes exact solutions via exponentiation eliminating integration error entirely.

**Structured matrices for stability.** AntisymmetricRNN constrains recurrent weights to be skew-symmetric for gradient stability [11]; Lipschitz RNNs extend this via symmetric-skew decompositions [12]; S4/Mamba use structured state matrices with HiPPO initialization for long-range sequence modeling [13], [14]. These methods use matrix structure for training stability or computational efficiency. LGN's S-D decomposition serves a different purpose, which is guaranteeing $\text{Re}(\lambda) \leq 0$ to encode physical stability and dissipation for system identification.

**Koopman methods.** Koopman theory linearizes nonlinear dynamics in lifted observable spaces [15]; DMD [16] and deep Koopman methods [17] learn such embeddings from data. These approaches target latent linearization of nonlinear systems; LGN learns the generator directly in native state space for systems that are already linear, where eigenvalues correspond to physical poles.

**Classical continuous-time identification.** Recovering $A$ from sampled trajectories involves the matrix logarithm: $F = \exp(A\Delta t)$ implies $A = \log(F)/\Delta t$. This inversion is non-unique at low sampling rates ("aliasing"), requiring structural priors for identifiability [18]. Hardt et al. [19] prove polynomial convergence guarantees for gradient descent on linear dynamical systems, establishing that unconstrained learning is tractable in principle. However, their analysis assumes exact gradients and does not address structural guarantees: the learned system may be unstable or non-dissipative even at convergence. Our 100D experiments show this, as unconstrained regression learns 16 unstable eigenvalues from stable data. LGN's S-D parameterization restricts optimization to the manifold of stable dissipative generators, providing spectral guarantees that unconstrained methods lack.

**Magnus expansion.** For time-varying systems $\dot{x} = A(t)x$, the Magnus expansion expresses solutions as $\exp(\Omega(t))$ with commutator corrections. Truncated Magnus integrators are a well-studied class of exponential/Lie-group methods that preserve qualitative structure at any truncation order. We leverage this for LTV systems, learning $A(t)$ from data rather than using Magnus solely as a solver for known dynamic.

# Conclusion

We introduced Lie Generator Networks (LGN), which learn structured generator matrices and evolve states via matrix exponentiation rather than numerical integration. Unlike Neural ODEs that learn black-box vector fields, or Hamiltonian networks restricted to conservative systems, LGN recovers the generator *A* whose eigenvalues directly encode stability, resonant frequencies, and damping rates. The key insight is that linearization, which ubiquitous in circuit analysis, control, and robotics, creates exactly the setting where LGN applies. By learning A(t) rather than f(x), and constraining A to structured forms, LGN guarantees physical invariants that integration-based methods violate. Correct eigenvalue recovery elevates the learned model to a system identification tool which enables understanding of the underlying physics of the system. The matrix A encodes decay timescales, resonant modes, and stability properties, all quantities that enable extrapolation, inverse design, and physical interpretation. This allows LGN-SD to go beyond interpolating trajectories to identifying the underlying system from measured, noisy data.

Future work includes automatic structure selection from data, extension to nonlinear systems via successive linearization, and application to inverse design where LGN's differentiability enables gradient-based optimization of system parameters.

# Limitations

**Representational scope.** The S−D decomposition guarantees stability in the Euclidean norm (P = I in the Lyapunov equation $A^\top P + PA \prec 0$), which is the natural metric for passive RLC networks where $\|x\|^2$ corresponds to stored energy. This excludes non-normal Hurwitz systems exhibiting transient growth as matrices with large off-diagonal entries can be stable yet not Euclidean-dissipative, and LGN-SD would suppress such entries to satisfy the symmetric negative-definite condition. Learning a Lyapunov metric P jointly with A [20] would extend coverage to all Hurwitz matrices without altering the Magnus machinery or training procedure.

**Scalability.** The matrix exponential costs $O(n^3)$ per evaluation, negligible at n = 100 but prohibitive beyond n ≈ 1000. Krylov subspace methods can approximate exp(A)v in $O(n^2 k)$ with k ≪ n [21], and exploiting the O(n)-sparse structure of physical networks (e.g., ladder topologies) would extend applicability to dimensions typical of device-level simulation (n = $10^3$–$10^5$), though we have not yet validated this empirically.

## Acknowledgements

This work was partially supported by the Department of Energy Grant No. DE-SC0022248, Office of Naval Research Grant No. N00014-21-1-2634, and Air Force Office of Scientific Research Grants No. FA9550-21-1-0305 and FA9550-22-1-0433.

# Appendix

## A.1 Magnus expansion

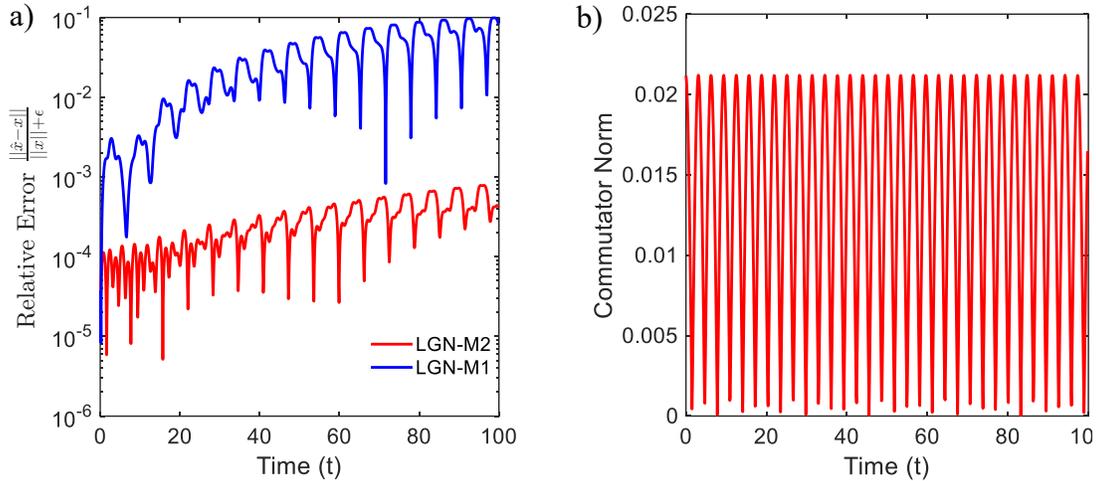

Figure A.1. **Magnus expansion effect.** a) LGN-M2 achieves 100× lower error than LGN-M1. b) Commutator norm $\|[A(t), A(t+\Delta t)]\|$ with mean 0.013.

When the parametric form of A(t) is known—here $\gamma(t) = \gamma_0(1 + \gamma_{\text{amp}}\sin(\omega_d t))$; LGN reduces to learning 4 scalar parameters. In this rigid setting, the commutator correction becomes critical: LGN-M2 outperforms LGN-M1 by ~100× (Fig. A.1). With no free parameters to absorb truncation error, second-order Magnus directly improves accuracy. This contrasts with the flexible S-D parameterization (main text), where the model compensates for first-order truncation by adjusting A(t), making M1 and M2 equivalent. More generally, commutator correction becomes essential when A(t) varies rapidly, Δt is large, or the system involves strongly non-commuting generators (e.g., rotating frames, coupled multi-physics systems).

## A.2 Comparison with Dissipative SymODEN

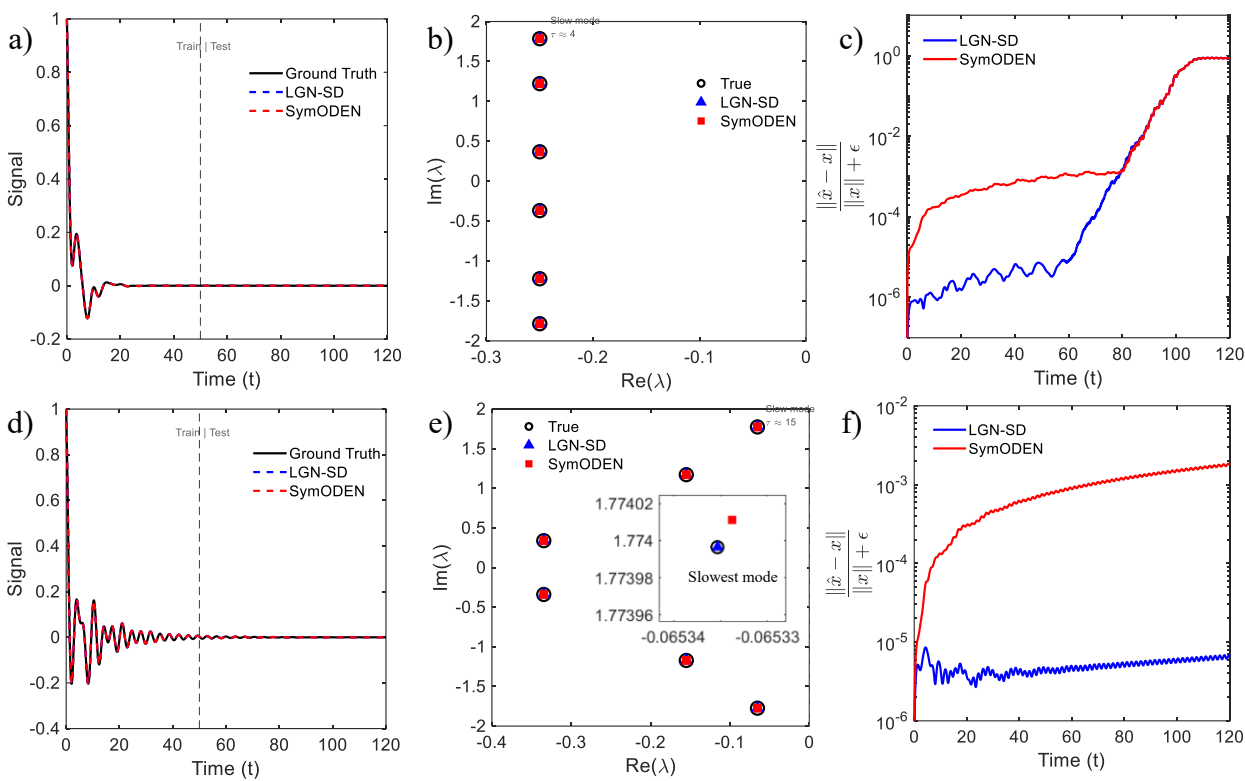

Figure A.2: **Stiff System Identification.** Top row: Uniform damping (R = [0.5, 0.5, 0.5], stiffness ratio = 1×). Bottom row: Stiff damping (R = [0.01, 0.1, 1.0], stiffness ratio ≈ 5×). (a,d) Trajectory of first state variable; dashed line marks the training horizon T = 50, by which all modes have fully decayed. (b,e) Eigenvalue recovery in the complex plane; the slowest mode ($\tau \approx 15$ in the stiff case) is shown in inset. (c,f) Relative prediction error over time. On the uniform system both methods perform comparably. On the stiff system, SymODEN's $\tau$-horizon training fails to recover the slow eigenvalue, producing two orders of magnitude higher relative error than LGN-SD.

We compare against Dissipative SymODEN which utilizes port-Hamiltonian dynamics with energy dissipation into the network architecture. For linear systems, we implement the corresponding parameterization: $A = (J - D)P$ with learnable skew-symmetric J (interconnection), positive semi-definite D (dissipation), and positive definite P (energy Hessian), following the port-Hamiltonian form $\dot{x} = (J - D)\nabla H$ with quadratic $H(x) = \frac{1}{2}x^T P x$. This is strictly favorable to the baseline: the optimizer searches a smaller space with no approximation error from neural networks. We further advantage the baseline by using L-BFGS-B with 5 random restarts (the original paper uses Adam), and setting τ = 5 (the original uses τ = 3). Both methods use identical training data (5 trajectories, T = 50, Δt = 0.1).

We construct a 6-dimensional RLC ladder (3 sections) in two configurations. The uniform case ($R_1 = R_2 = R_3 = 0.5$) produces eigenvalues with similar real parts, giving a stiffness ratio of 1×. The stiff case ($R_1 = 0.01$, $R_2 = 0.1$, $R_3 = 1.0$) produces eigenvalues spanning a factor of ~5×, with the slowest mode having time constant τ ≈ 15. Figure A.2 (top row) confirms that on the uniform system, both methods achieve comparable accuracy — eigenvalues overlap, trajectories are visually indistinguishable, and relative errors converge to similar levels. This serves as a control validating that neither method is artificially advantaged. On the stiff system (bottom row), a clear separation emerges. LGN-SD recovers all six eigenvalues to high accuracy, including the slow mode at Re(λ) ≈ −0.065 (inset, Fig. A.2e). Dissipative SymODEN recovers the fast modes but misses the slow eigenvalue entirely. The relative error reflects this: LGN-SD maintains ~$10^{-5}$ while Dissipative SymODEN plateaus at ~$10^{-3}$, a gap of two orders of magnitude.

The failure mechanism is intrinsic to τ-horizon training, not to the port-Hamiltonian parameterization itself. Each training window spans τΔt = 0.5 time units, during which the slow mode (τ ≈ 15) completes only ~3% of its decay — indistinguishable from a constant offset within the window. The full training horizon T = 50 spans over three time constants of the slowest mode, meaning the data contains the complete transient response — but the τ-window training procedure cannot access it. LGN-SD avoids this entirely: the matrix exponential encodes all timescales simultaneously, and full-trajectory fitting ensures that even the slowest mode contributes a large, unambiguous signal to the loss function.